\documentclass[review,12pt,authoryear, preprint]{elsarticle}

\usepackage{amssymb}

\usepackage{color}
\usepackage{xcolor}
\usepackage{subfigure}
\usepackage{amsmath,amsfonts}
\usepackage{url}
\usepackage{booktabs}
\definecolor{cvprblue}{rgb}{0.21,0.49,0.74}
\usepackage[breaklinks,colorlinks,citecolor=cvprblue]{hyperref}

\usepackage{booktabs}
\usepackage{amssymb}
\usepackage{amsmath}
\usepackage{float}
\usepackage{bm}
\usepackage{stfloats}
\usepackage{multirow}
\usepackage{mathrsfs}
\usepackage{algorithmic}
\usepackage[ruled,linesnumbered]{algorithm2e}
\usepackage{colortbl}

\journal{Elsevier}

\begin{document}
	
	\begin{frontmatter}
		
		\title{Open-Pose 3D Zero-Shot Learning: Benchmark and Challenges}

		\author[1,2]{Weiguang Zhao\fnref{cor2}}
		\ead{Weiguang.Zhao@liverpool.ac.uk}
		
		\author[6]{Guanyu Yang\fnref{cor2}} 
		\ead{Guanyu.Yang@dukekunshan.edu.cn}
		
		\author[2]{Rui Zhang\corref{cor1}}
		\ead{Rui.Zhang02@xjtlu.edu.cn}
		
		\author[6]{Chenru Jiang} 
		\ead{Chenru.Jiang@dukekunshan.edu.cn}
		
		\author[1,3]{Chaolong Yang}
		\ead{Chaolong.Yang@liverpool.ac.uk}
		
		\author[4]{Yuyao Yan}
		\ead{yuyao.yan@xjtlu.edu.cn}
		
		\author[5]{Amir Hussain}
		\ead{A.Hussain@napier.ac.uk}
		
		\author[6]{Kaizhu Huang\corref{cor1}} 
		\ead{Kaizhu.Huang@dukekunshan.edu.cn}

		\address[1]{Department of Computer Science, University of Liverpool, Liverpool L69 7ZX, UK.}
		\address[2]{Department of Foundational Mathematics, Xi'an Jiaotong-Liverpool University, Suzhou, 215123, China.}
		\address[3]{Department of Mechatronics and Robotics, Xi'an Jiaotong-Liverpool University, Suzhou, 215123, China.}
		\address[4]{School of Robotic, Xi'an Jiaotong-Liverpool University, Suzhou, 215123, China.}
		\address[5]{School of Computing, Edinburgh Napier University, Edinburgh, EH11 4BN, UK.}
		\address[6]{Data Science Research Center, Duke Kunshan University, Kunshan, 215316, China.}
		
		\cortext[cor1]{Corresponding authors}
		\fntext[fn1]{Equal contribution}

		\begin{abstract}
			
			With the explosive 3D data growth, the urgency of utilizing zero-shot learning to facilitate data labeling becomes evident. Recently, methods transferring language or language-image pre-training models like Contrastive Language-Image Pre-training (CLIP) to 3D vision have made significant progress in the 3D zero-shot classification task. These methods primarily focus on 3D object classification with an aligned pose; such a setting is, however, rather restrictive, which overlooks the recognition of 3D objects with open poses typically encountered in real-world scenarios, such as an overturned chair or a lying teddy bear. To this end, we propose a more realistic and challenging scenario named open-pose 3D zero-shot classification, focusing on the recognition of 3D objects regardless of their orientation. First, we revisit the current research on 3D zero-shot classification, and propose two benchmark datasets specifically designed for the open-pose setting. We empirically validate many of the most popular methods in the proposed open-pose benchmark.  Our investigations reveal that most current 3D zero-shot classification models suffer from poor performance, indicating a substantial exploration room towards the new direction. Furthermore, we study a concise pipeline with an iterative angle refinement mechanism that automatically optimizes one ideal angle to classify these open-pose 3D objects. In particular, to make validation more compelling and not just limited to existing CLIP-based methods, we also pioneer the exploration of knowledge transfer based on Diffusion models. While the proposed solutions can serve as a new benchmark for open-pose 3D zero-shot classification, we discuss the complexities and challenges of this scenario that remain for further research development. The code is available publicly at \href{https://github.com/weiguangzhao/Diff-OP3D}{\tt\small\text{https://github.com/weiguangzhao/Diff-OP3D}}.
			
		\end{abstract}

		
	\end{frontmatter}

	\section{Introduction}
	\label{sec:intro}
	Deep learning models have achieved remarkable advancements in computer vision tasks~\citep{xue2023ulip, chen2023viewnet, zhao2023divide, jiang2023pointgs}. However, their outstanding performance relies heavily on large amounts of labeled data. In this regard, zero-shot learning, where classes are learned without corresponding samples, has drawn substantial scholarly focus. While 2D zero-shot classification research~\citep{radford2021learning, wang2019survey, naeem2023i2mvformer, ye2023rebalanced, ye2021disentangling} is thriving, research on 3D zero-shot classification~\citep{ZhangGZLM0QG022,Zhu2022PointCLIPV2,naeem20223d} is still in nascent stages. Considering the inherent irregularity and sparsity of 3D data, the extracted features exhibit significant differences compared to 2D data. As a result, most existing 2D zero-shot learning methods fail to produce effective results when applied directly to the 3D domain~\citep{cheraghian2019zero, CheraghianRCP19, cheraghian2020transductive, cheraghian2022zero}.
	
	\begin{figure}[h]
		\centering
		\includegraphics[width=0.99\textwidth]{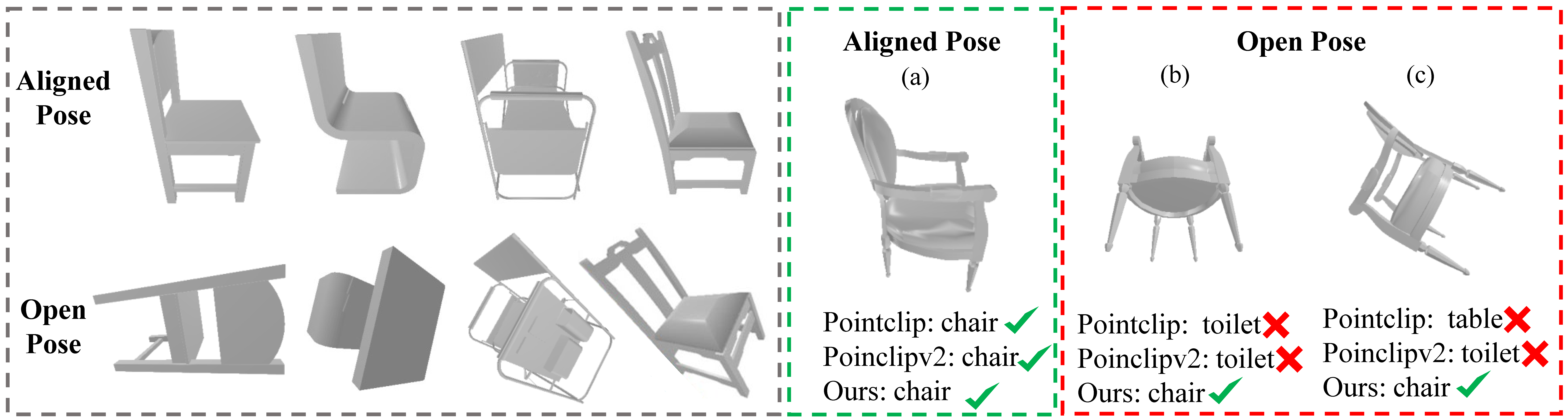}
		\caption{3D Zero-Shot Classification for Aligned-Poses and Open-Poses. (a) is a 3D sample in aligned-pose from the dataset ModelNet40, while (b) and (c) are the corresponding sample in open-poses from our benchmark ModelNet40$^{\ddagger}$. }
		\label{fig:intro}
	\end{figure}
	
	Currently, to better leverage knowledge from the 2D domain, most state-of-the-art methods (SOTAs)~\citep{ZhangGZLM0QG022, huang2023clip2point, Zhu2022PointCLIPV2, xue2023ulip, wang2023transferring} explore bridging Contrastive Language-Image Pre-training (CLIP)~\citep{radford2021learning} to 3D classification. Their general pipeline is to project 3D objects to 2D depth maps and then to match the class names and image features via CLIP. As an initial exploration, these methods have already demonstrated remarkable effectiveness on existing benchmarks. However, all these evaluations are based on the assumption that 3D objects are in aligned-poses. Such alignment is based on prior category knowledge, thus inherently simplifying the task of classification. Unfortunately, objects in the real world are typically positioned in random orientations, which can be referred to as open-pose, such as an overturned chair or a flying bird. These open-pose scenarios present a realistic yet significant challenge for current SOTAs. As shown in Fig.~\ref{fig:intro}, the effectiveness of these SOTAs is closely tied to the objects' orientations, and their performance becomes poor when the aligned 3D object is modified to the open-pose case.
	
	To this end, we investigate a more challenging, yet practical task of 3D zero-shot classification aiming at recognizing 3D objects in open-poses. Despite its importance, this open-pose setting is rarely studied in the literature. To this end, we first revisit the current methods and datasets for 3D zero-shot classification.  Furthermore, we propose to generate two  new datasets to benchmark open-pose 3D zero-shot classification, by applying random rotations to each sample in the widely-used zero-shot datasets ModelNet40~\citep{wu20153d} and McGill~\citep{siddiqi2008retrieving}. Due to the uncertain orientations of objects, the selection of projection angles becomes a crucial matter. In this context, we propose a concise pipeline with an iterative angle refinement mechanism that automatically optimizes one ideal project angle to classify these open-pose 3D objects. Specifically, this mechanism dynamically determines the projection angles based on the characteristics of the input object and the outcomes of matching processes which shows potential in handling open-pose variations.
	
	Additionally, it is notable that existing projection-based approaches exclusively use CLIP as their text-image matching backbone, and their projections are primarily limited to depth maps and rendered images. To achieve more comprehensive evaluations, we make the first attempt to utilize a 2D pre-trained diffusion model~\citep{rombach2022high, sohl2015deep} as the backbone for classifying 3D objects. We also incorporate the edge image, thereby expanding the variety of projection styles. 
	
	While the above pipeline can serve as the first as well as a new benchmark for open-pose 3D zero-shot classification, we also discuss the challenges and potential outlooks for 3D zero-shot classification in the open-pose setting, which may further inspire future work in this direction.
	
	The contributions of our work can be summarized as follows:
	
	\begin{itemize}
		\item We propose a more challenging scenario, namely open-pose 3D zero-shot classification, to uncover the current limitations of state-of-the-art approaches. We survey a comprehensive set of 3D zero-shot classifications in the open-pose setting for the first time.
		
		\item We develop two benchmarks, ModelNet40$^{\ddagger}$ and McGill$^{\ddagger}$ for evaluating open-pose 3D zero-shot classification.  Our empirical investigations reveal that  most current SOTA 3D zero-shot methods suffer from poor performance for open-pose classification. 
		
		\item We design a concise pipeline with an iterative angle refinement mechanism, which achieves a substantial improvement, thus presenting a new benchmark method in the open-pose setting.
		
		\item We discuss the challenges for open-pose 3D zero-shot classification, and set out new directions which may inspire future research towards this direction.

	\end{itemize}
	
	\section{Related Work}
	\label{secR}
	
	\textbf{Classification in the 3D Domain:} The mainstream deep learning methods for 3D classification can be categorized into three types~\citep{guo2020deep}: point-based, voxel-based, and multi-view-based methods. PointNet~\citep{qi2017pointnet} was the pioneering network to extract point cloud features and recognize 3D objects. Point-Transformer~\citep{zhao2021point, wu2022point} utilizes transformer architecture in the 3D domain, leading to a significant improvement in classification performance. Additionally, Voxnet~\citep{maturana2015voxnet} introduces voxel representation to process point clouds, while MinkowskiNet~\citep{choy20194d} offers a sparse voxel convolution architecture based on this concept. Moreover, MVCNN~\citep{su2015multi} suggests projecting 3D mesh categories from multiple views and employing 2D networks for training. LRMV~\citep{yang2019learning} predicts the score of each projection view to select images. Some studies~\citep{rotationnet, wei2021learning, zhou2024msdcnn} have also delved into how viewpoint selection can influence classification results. Unlike these supervised learning methods, our work is focused on zero-shot classification, where new (unseen) classes emerge during testing.
	
	\textbf{Zero-Shot Classification in the 2D Domain:} The objective of conventional zero-shot learning tasks is to classify a sample set from unseen categories~\citep{larochelle2008zero}. Presently, there are two primary approaches to addressing zero-shot classification tasks in the 2D domain~\citep{yang2022comprehensive}: embedding methods~\citep{RahmanKP20, zhou2022information, HanFCY22, ye2023rebalanced}, and generative methods~\citep{li2023your, YangLLW23}. The former seeks to learn a specific feature space that aligns with both the samples and category descriptions, while the latter transforms the zero-shot classification task into a standard classification problem by generating pseudo samples for unseen categories.
	
	Furthermore, both CLIP~\citep{radford2021learning} and the diffusion classifier~\citep{li2023your} can perform classification tasks without relying on the training set of the specific benchmark. Consequently, these methods are also considered zero-shot models. However, it's important to note that their pre-training dataset may include samples corresponding to classes within the target dataset. Therefore, strictly speaking, these methods differ from conventional zero-shot models in classifying ``unseen classes". Nonetheless, when applied to assist in the zero-shot classification of 3D data, they can be accurately categorized as zero-shot methods, as no 3D samples are used during the training process.
	
	\section{Revisiting 3D Zero-Shot Classification }
	In the realm of 3D zero-shot classification, two broad methods have emerged: language pre-training model-based and language-image pre-training model-based. The former relies on Word2Vec~\citep{mikolov2013distributed} or GloVe~\citep{pennington2014glove} to generate text embeddings for categories, while the latter primarily uses CLIP to match text embeddings with projected image embeddings. While Cheraghian et al.~\citep{cheraghian2022zero} provided a comprehensive review of the language pre-training model-based method, recent years have witnessed a surge in new 3D zero-shot classification approaches, particularly those leveraging the language-image pre-training model-based paradigm. In this section, we'll delve into a detailed overview and analysis of existing 3D zero-shot datasets and methodologies.
	
	\subsection{3D Zero-Shot Classification Benchmark Datasets.} 
	\label{3zcd}
	The existing 3D zero-shot classification benchmark datasets primarily consist of ModelNet40~\citep{wu20153d}, ModelNet10~\citep{wu20153d}, McGill~\citep{siddiqi2008retrieving}, ScanObjectNN~\citep{scanobjectnn}, and SHREC2015 \citep{SHREC2015}. In Table~\ref{tab:dataset}, we summarize the division of these datasets into seen and unseen classes, as well as their partitioning into training, validation, and test sets.

	\begin{table}[h]
		\resizebox{0.99\textwidth}{!}{
			\begin{tabular}{ccccc}
				\bottomrule
				Datasets           & Total Classes & Seen/Unseen Classes & Train/Valid/Test Samples \\ \hline
				ModelNet40       & 40            & 30/-                & 5852/1560/-              \\
				ModelNet10       & 10            & -/10                & -/-/908                  \\
				McGill            & 19            & -/14                & -/-/115                  \\
				ScanObjectNN     & 15            & -/11                & -/-/495                  \\
				SHREC2015         & 50            & -/30                & -/-/192                  \\ \bottomrule
			\end{tabular}
		}
		\caption{3D Zero-Shot Classification Benchmark Datasets.}
		\label{tab:dataset}
	\end{table}

	\noindent \textbf{ModelNet40 \& ModelNet10}: ModelNet40 is widely used in various 3D classification tasks, including 3D full-supervision, weakly-supervision, and few-shot classification. It consists of 40 categories of 3D objects, with each category containing a varying number of CAD models. The distribution of models across categories is balanced. This dataset encompasses a diverse array of common object classes, including chairs, tables, airplanes, cars, and more. Furthermore, ModelNet10 is a subset of ModelNet40, comprising only 10 categories of 3D objects. Most existing 3D zero-shot classification methods~\citep{narayan2020latent,michele2021generative,cheraghian2022zero,Hao2023ContrastiveGN} treat the 30 classes of ModelNet40 as seen classes and the remaining 10 classes as unseen classes, namely ModelNet10. It's worth noting that some language-image pre-training model-based methods~\citep{xue2023ulip,Zhu2022PointCLIPV2, qi2023recon, huang2023clip2point} consider all classes as unseen classes, as they directly utilize language-image pre-training models for inference without the need for training on seen classes.
	
	\noindent \textbf{McGill}: McGill dataset differs from ModelNet40 in its composition, focusing on various 3D biological samples such as fish, dinosaurs, spiders, octopuses, and more. Specifically, the McGill dataset comprises a total of 19 categories, with five categories overlapping with the visible classes in ModelNet40. Consequently, only the remaining 14 categories are utilized as unseen classes. Due to its high-quality point clouds and a wide range of object categories, this dataset is widely adopted for numerous 3D zero-shot classification methods~\citep{narayan2020latent,michele2021generative,cheraghian2022zero,Hao2023ContrastiveGN}.
	
	\noindent  \textbf{ScanObjectNN}: ScanObjectNN dataset comprises 15 categories of 3D objects derived from real-world scans, often presenting cluttered backgrounds and partial objects due to occlusions. These objects predominantly stem from indoor scenes. With four classes overlapping with ModelNet40 excluded, the remaining 11 classes are regarded as unseen classes. Due to varying qualities of point clouds and backgrounds, this dataset is frequently divided into three subsets: OBJ\_ONLY, OBJ\_BG, and PB\_T50\_RS.

	\noindent \textbf{SHREC2015}: This dataset was initially introduced at the Eurographics workshop for a 3D object retrieval competition. It comprises a total of 50 categories, with 20 of these categories overlapping with the seen classes in ModelNet40. Consequently, the remaining 30 categories are utilized as unseen classes. Moreover, SHREC2015 remains relatively underutilized in the zero-shot classification domain, with only three studies~\cite{cheraghian2019zero,cheraghian2020transductive, CheraghianRCP19} conducted for validation on this dataset.

	However, most of these datasets were deliberately aligned or oriented during collection. This makes 3D zero-shot classification tasks detached from complex real-world scenarios, where objects may be in arbitrary poses. In light of this, we make the first attempt to build an open-pose benchmark where all unseen 3D objects are in arbitrary poses.
	
	Taking into account the coverage and frequency of dataset usage, we decided to separately utilize the ModelNet40 and McGill datasets to create the open-pose datasets, referred to as ModelNet40$^{\ddagger}$ and McGill$^{\ddagger}$. We rotate each sample in ModelNet40 and McGill to obtain the ModelNet40$^{\ddagger}$ and McGill$^{\ddagger}$ datasets, respectively\footnote{The random angles and open-pose datasets are available on our GitHub.}. Although the samples in McGill inherently have less standardized orientation compared to those in ModelNet40, the range of their rotational angles is limited, lacking sufficient randomness to effectively illustrate the open-pose issue. Therefore, we also apply random angle rotations to them. Similar to previous datasets~\citep{wu20153d, siddiqi2008retrieving}, we divide our open-pose datasets as shown in Tab.~\ref{tab:dataset_op}.

	\begin{table}[h]
		\resizebox{0.95\textwidth}{!}{
			\begin{tabular}{ccccc}
				\bottomrule
				Datasets           & Total Classes & Seen/Unseen Classes & Train/Valid/Test Samples \\ \hline
				ModelNet40$^{\ddagger}$         & 40            & 30/-                & 5852/1560/-              \\
				ModelNet10$^{\ddagger}$          & 10            & -/10                & -/-/908                  \\
				McGill$^{\ddagger}$            & 19            & -/14                & -/-/115                  \\
				\bottomrule
			\end{tabular}
		}
		\caption{ Our Open-Pose 3D Zero-Shot Classification Benchmark Datasets. }
		\label{tab:dataset_op}
	\end{table}
	
	\subsection{3D Zero-Shot Classification with Language Pre-training Model} 
	The language pre-training model-based methods draw inspiration from existing 2D zero-shot approaches, using text embeddings to establish correlations between seen and unseen classes. Cheraghian et al.~\citep{CheraghianRCP19,cheraghian2019zero,cheraghian2020transductive,cheraghian2022zero} pioneered research on 3D zero-shot classification. They made the first attempt in traditional~\citep{cheraghian2019zero}, inductive~\citep{CheraghianRCP19}, and transductive~\citep{cheraghian2020transductive} zero-shot classification settings within the 3D domain, providing a standardized evaluation protocol for subsequent research work. Their proposed methods are based on text embeddings (Word2Vec~\citep{mikolov2013distributed} \& GloVe~\citep{pennington2014glove}) with two refinement loss functions~\citep{cheraghian2022zero}. On the other hand, 3DCZSL~\citep{naeem20223d} takes an approach that considers the 3D zero-shot classification task from the perspective of geometric structure composition. However, it relies on point-wise component labels, which currently can only be satisfied by the PartNet dataset~\citep{Mo_2019_CVPR}. 3DGenZ~\citep{michele2021generative} is proposed as the first generative 3D zero-shot classification network. It conducts Google searches for 100 images per class and utilizes a pre-training classification network to obtain image representations for each category.
	
	\subsection{3D Zero-Shot Classification with  Language-Image Pre-training Model} 
	We make the first attempt to provide an overview of the language image pre-training model-based methods. Currently, all these methods utilize CLIP's knowledge for zero-shot classification of 3D models. Based on their implementation approaches, we categorize them into two groups: input-optimization and encoder-distillation methods.
	
\begin{figure}[h]
		\centering
		\includegraphics[width=0.75\textwidth]{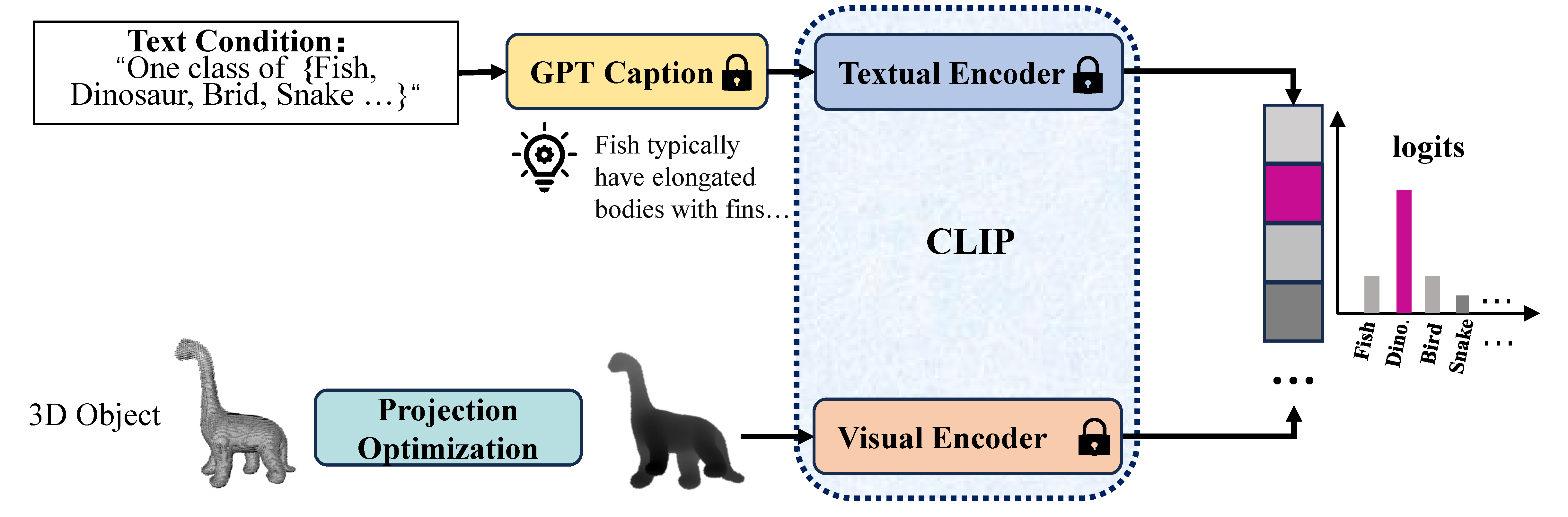} 
		\caption{Input-optimization Framework}
		\label{fig:iof}
	\end{figure}

	\textbf{Input-optimization methods.} These methods typically involve optimizing the text input of CLIP or rendering image inputs, as depicted in Fig.~\ref{fig:iof}. PointCLIP~\citep{ZhangGZLM0QG022} pioneered the projection of 3D objects into 2D images and utilized CLIP for category matching. Building upon this work, PointCLIPv2~\citep{Zhu2022PointCLIPV2} incorporates large-scale language models (such as GPT-3)~\citep{BrownMRSKDNSSAA20} to automatically design more descriptive 3D-semantic prompts. Furthermore, DiffCLIP~\citep{diffclip} proposes integrating stable diffusion with ControlNet~\citep{controlnet} to minimize the domain gap between rendered images and realistic images. Additionally, DILF~\citep{dilf} utilizes GPT-3 to generate textual prompts enriched with 3D semantics and designs a differentiable renderer with learnable rendering parameters to produce representative multi-view images. Most of these methods do not require additional training and perform direct inference using existing pre-trained models. However, they often require a considerable number of hyperparameters to achieve the best results.
	
\begin{figure}[h]
		\centering
		\includegraphics[width=0.7\textwidth]{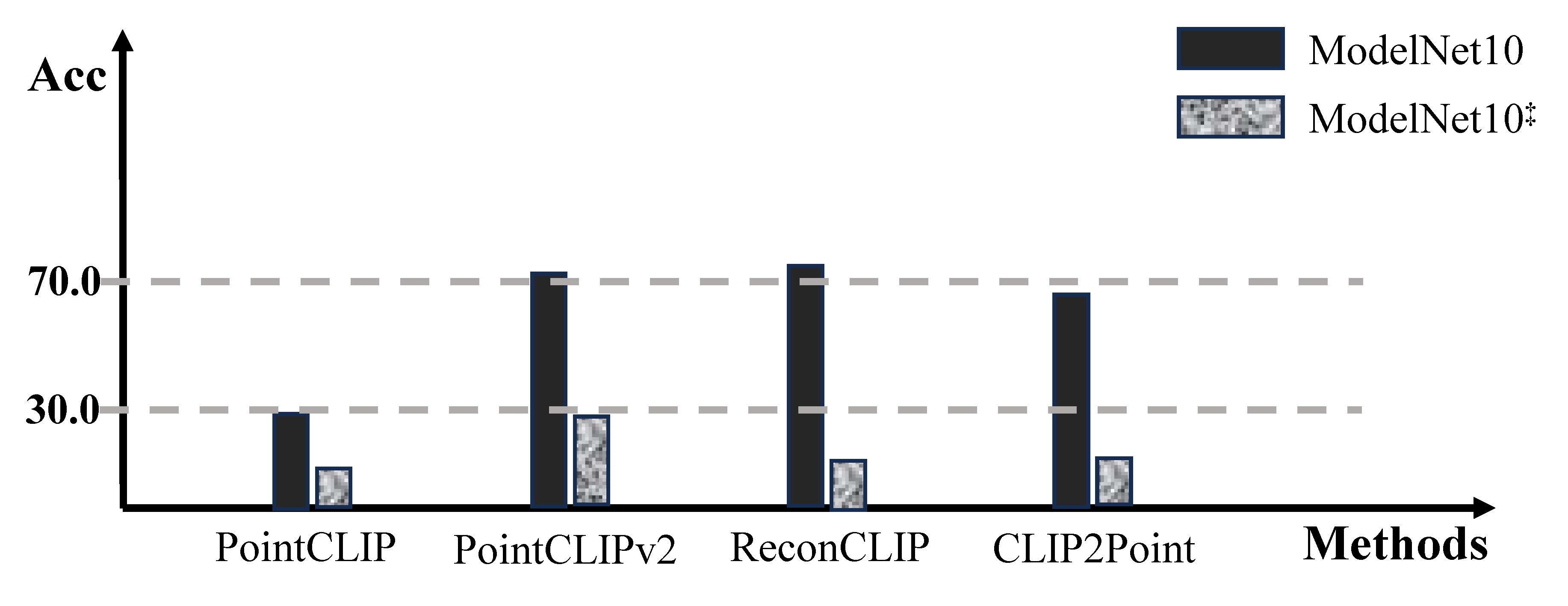} 
		\caption{Performance on Aligned and Open-Pose Dataset}
		\label{fig:paod}
	\end{figure}

	\textbf{Encoder-distillation methods.} As depicted in Fig.~\ref{fig:paod}, these methods retain the text encoder of CLIP while incorporating a visual encoder to supervise their newly designed encoder. Specifically, Ulip~\citep{xue2023ulip, xue2023ulipv2} and CLIPgoes3D~\citep{CLIPgoes3D} design a new 3D encoder to learn the relationship between 3D features, rendered images, and textual descriptions. Furthermore, ReconCLIP~\citep{qi2023recon} leverages reconstruction tasks to enhance the feature extraction of 3D encoders. Moreover, CLIP2Point~\citep{huang2023clip2point} develops a new image encoder network to narrow the domain gap between depth maps and realistic images. Methods of this kind often require additional 3D data for pre-training, limiting their applicability to the domain of the training data. Furthermore, they heavily rely on 3D data, resulting in significant computational overhead. For instance, Ulip and CLIPgoes3D both require training on 8 A100 GPUs.
	
	\begin{figure}[h]
		\centering
		\includegraphics[width=0.7\textwidth]{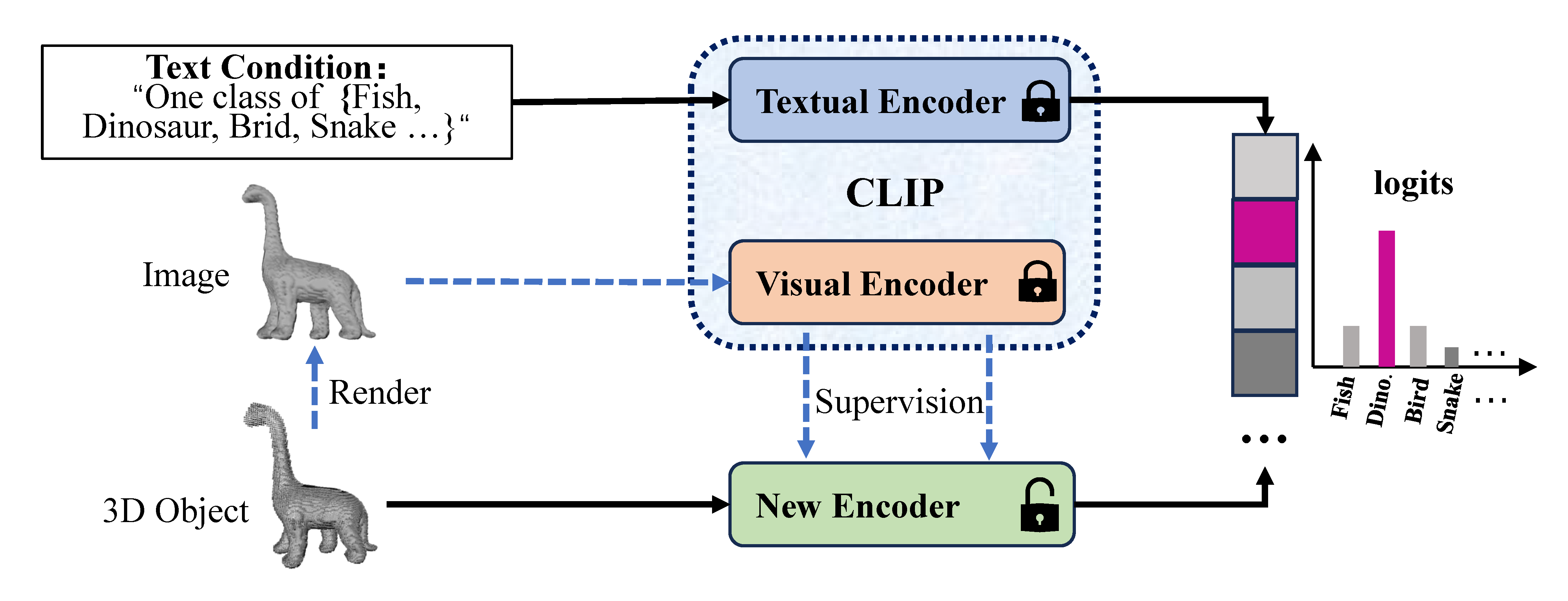} 
		\caption{Encoder-distillation Framework}
		\label{fig:edf}
	\end{figure}

	However, both input-optimization and encoder-distillation methods are built upon the assumption that objects are in aligned poses. As depicted in Fig.~\ref{fig:paod}, the more practical and challenging task of recognizing 3D objects in open poses is neglected. In our work, we explore the performance of many existing methods in an open-pose setting and propose a new method tailored specifically for this scenario as a baseline.
	
	\section{Our Main Methodology}
	\label{secM}
\begin{figure}[h]
		\centering
		\includegraphics[width=0.99\textwidth]{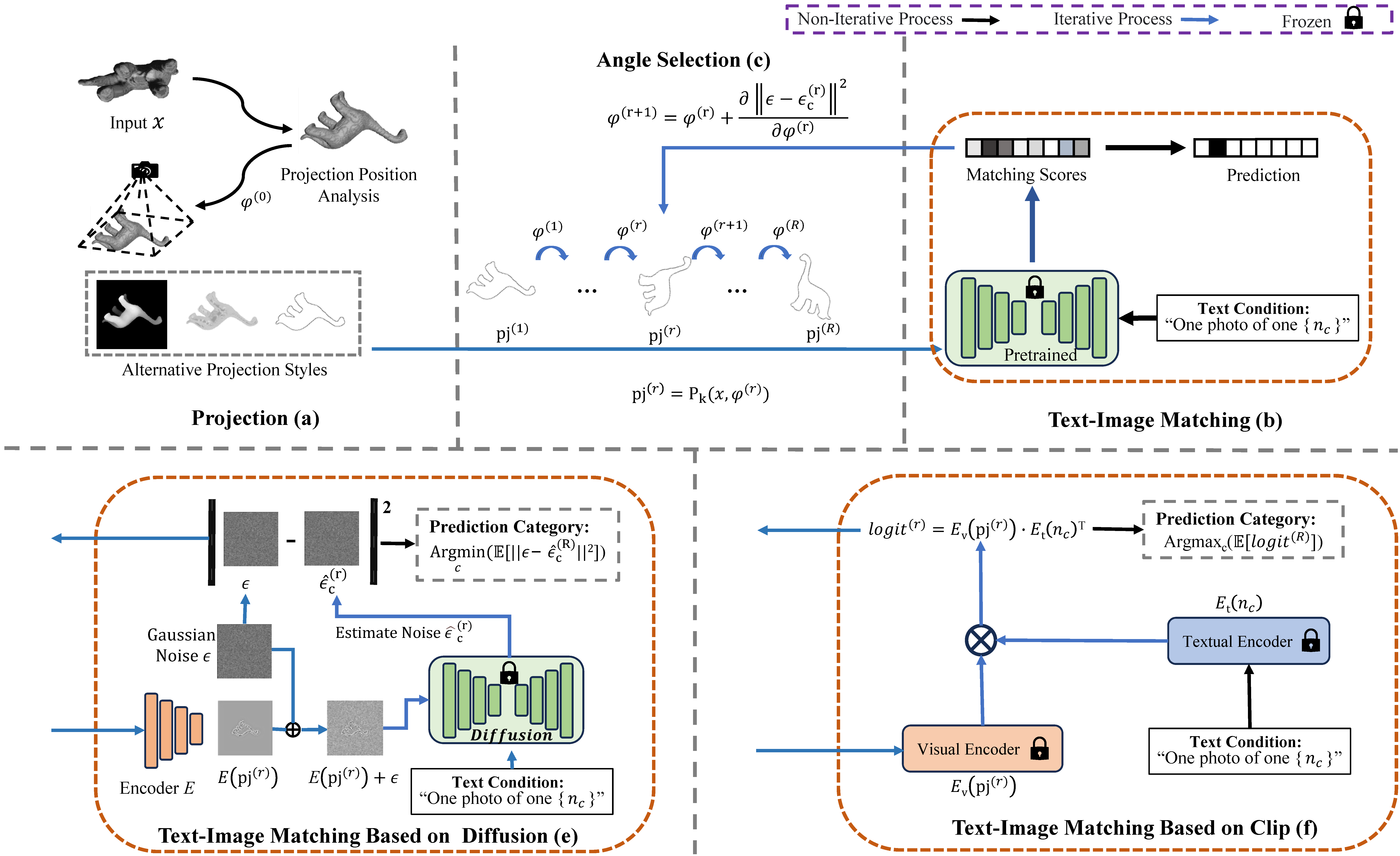} 
		\caption{Overview of Our Pipeline}
		\label{fig:pipeline}
	\end{figure}

	As depicted in Fig.~\ref{fig:pipeline}, our pipeline consists of three main components: Projection (a), Text-Image Matching (b), and Angle Selection (c). Given a 3D object, projection images can be obtained by selecting projection angles and styles. These images are then matched with text descriptions via a pre-training text-image matching backbone. Additionally, the projection angles can be pre-selected or determined based on the input object and matching outcomes. With the final settled projection angles, predictions are obtained based on the corresponding matching scores. Furthermore, (e) and (f) are optional text-image matching backbones based on Diffusion and CLIP, respectively. In this section, we provide detailed descriptions of this pipeline.
	
	\subsection{Projection}
	We utilize a perspective projection to transform the 3D input $x$ into 2D images. In the projection phase, our main focus lies on two aspects: projection position analysis and projection style selection. Here, we concentrate on exploring various styles for projection. The analysis of projection positions, closely linked with angle selection, will be elaborated in Section~\ref{AS}.
	
\begin{figure}[h]
		\centering
		\includegraphics[width=0.6\textwidth]{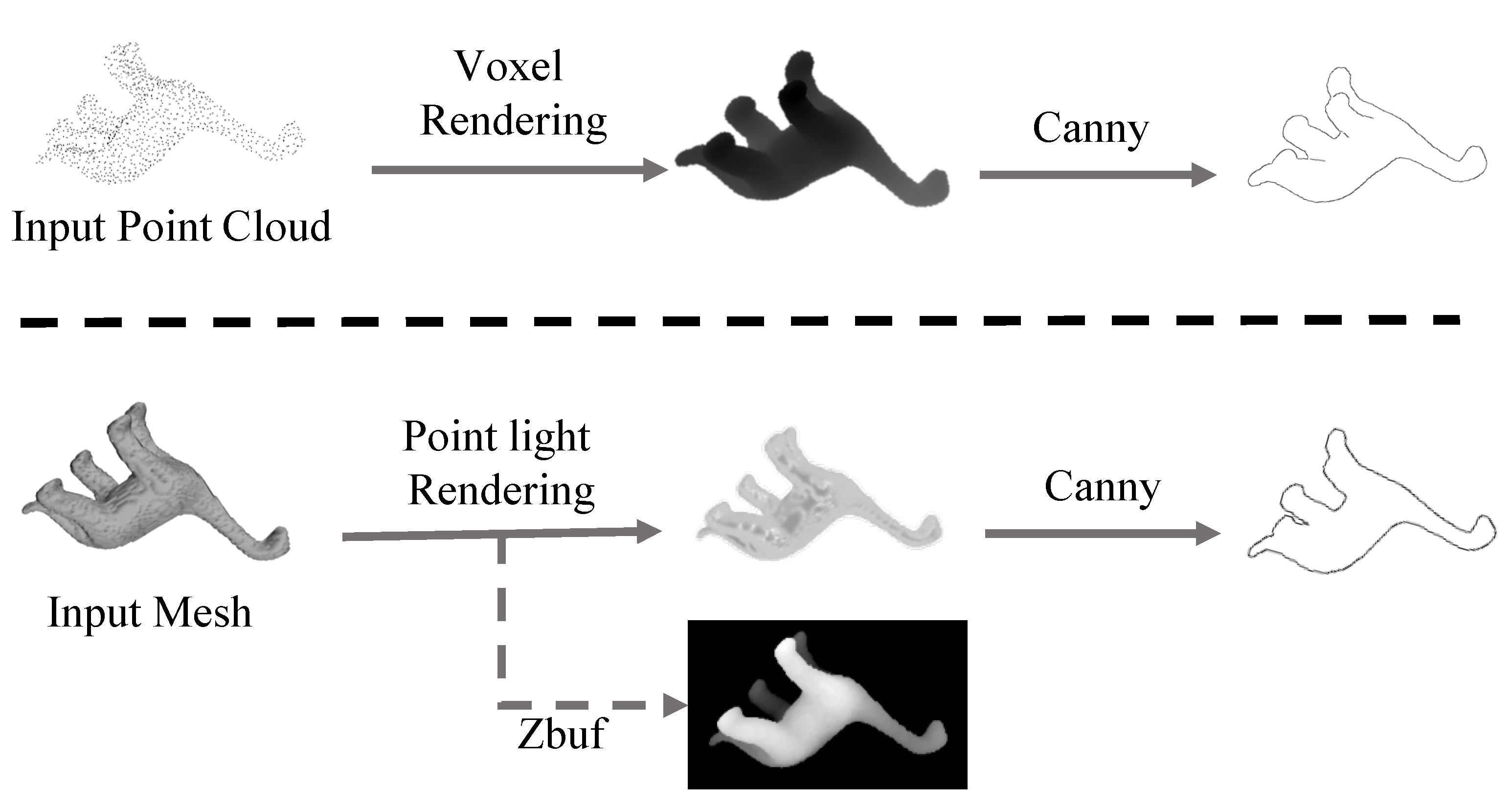}
		\caption{Projection Styles}
		\label{fig:projection}
	\end{figure}

	In addition to the depth maps and rendered images commonly used in existing works, we introduce edge maps as an additional projection style. Given that the primary data representations for 3D objects are point clouds and meshes, our method considers both inputs. Due to the sparsity of point clouds, their projection style differs significantly from that of meshes. Hence, we convert images projected from both point clouds and meshes into edge images for zero-shot classification. As depicted in Fig.~\ref{fig:projection}, the two different input data formats yield distinct projected images, which are ultimately extracted into similar edge images.
	
	Specifically, when the input is point-cloud data, direct projection yields a set of discrete pixels, making it challenging to adequately represent its semantic information. Following the approach of PointCLIPv2~\citep{Zhu2022PointCLIPV2}, we adopt voxel projection, which involves four steps: voxelization, densification, smoothing, and squeezing, to obtain a continuous pixel projection image. Furthermore, we utilize the Canny algorithm to compute pixel gradients, thereby obtaining the edge image. On the other hand, mesh data includes additional face information, specifically triangular surfaces, compared to point-cloud data. In this regard, we can directly achieve the continuous pixel projection image by point light projecting. To simulate parallel light, we position the point light source far away from the camera and the 3D mesh. To address noise on the surface of mesh data, we increase the Canny gradient threshold~\citep{canny} to mitigate this impact, which also results in the removal of some detailed information. For instance, the edge image of the mesh lacks the boundary of the legs compared to the edge image of the point cloud in Fig.~\ref{fig:projection}.

	\subsection{Text-Image Matching}
	\label{sec:DC}
	Since CLIP is originally designed for text-image matching, its performance in 3D zero-shot classification tasks has been extensively explored in numerous studies. Conversely, the diffusion process has not yet been applied to this area. Therefore, we revisit how the diffusion process, as a generative model, accomplishes this task.
	
	Given a clean feature sample $\bm{f_{0}}$ and a variance schedule $\beta_{1},...,\beta_{T}$, we can define a Markov chain that gradually adds Gaussian noise to the data as follows: 
	\begin{equation}
		q(\bm{f_{t}}\vert \bm{f_{t-1}})=\mathcal{N}(\bm{f_{t}}; \sqrt{1-\beta_{t}}\bm{f_{t-1}},\beta_{t}\bm{I}).
	\end{equation}
	
	The denoising diffusion probabilistic models~\citep{rombach2022high} learn the reverse process $p_{\theta}(\bm{f_{t-1}}\vert \bm{f_{t}}, s)$ with the corresponding semantic description $s$. Typically, the diffusion model can be interpreted as a denoising autoencoder $\bm{\epsilon _{\theta}}(\bm{f_{t}}, t, s)$ with a training target that minimizes the objective function:
	\begin{equation}
		L_{DM} = \mathbb{E}_{\bm{f_{0}},\bm{\epsilon} \sim \mathcal{N}(\bm{0},\bm{I}),t}\left[\Vert\bm{\epsilon} - \bm{\epsilon _{\theta}}(\bm{f_{t}},t,s)\Vert^{2}_{2}\right].
	\end{equation}
	
	In the Diffusion Classifier~\citep{li2023your}, the posterior probability of the latent feature of the sample conditioned on the specific semantic is calculated based on the denoising performance, thus enabling the classification task. With a semantic description constructor $S(c)$ for each class and a 2D feature extractor $\bm{\epsilon _{\theta}}()$, such probability can be approximately estimated as follows:
	\begin{equation}
		p_{\theta}(c\vert \bm{f_{0}}) = \frac{\exp\left\{ - \mathbb{E}_{t,\bm{\epsilon}}\left[\Vert \bm{\epsilon} -\bm{\epsilon _{\theta}}(\bm{f_{t}},t,S(c)) \Vert^{2}_{2}\right] \right\}}{\sum_{j}\exp\left\{ - \mathbb{E}_{t,\bm{\epsilon}}\left[\Vert \bm{\epsilon} -\bm{\epsilon _{\theta}}(\bm{f_{t}},t,S(j)) \Vert^{2}_{2}\right] \right\}}.
	\end{equation}
	Here, we directly adopt such a pre-training diffusion framework with the encoder $E(\cdot)$. Denoting the multi-style projection process as the function $P_{k}(\cdot,\bm{\varphi})$ with style $k \in K$ and angles $\bm{\varphi}$, we define the text-image matching score with the latent denoising module as below:
	\begin{equation}
		\label{eq:MD}
		\begin{aligned}
			{M\! S}_{\bm{\theta}, K}(\bm{x}, \bm{\varphi}, c) &= \exp\left\{ - \mathbb{E}_{t,\bm{\epsilon},k}\left[\Vert \bm{\epsilon} -\bm{\hat{\epsilon}}_{\bm{\theta},k}(\bm{x},\bm{\varphi},c) \Vert^{2}_{2}\right] \right\}\\
			\bm{\hat{\epsilon}}_{\bm{\theta},k}(\bm{x},\bm{\varphi},c) &= \bm{\epsilon_{\theta}}\left(\bm{f}_{t,k}\left(\bm{x}, \bm{\varphi}\right),t,S_{k}\left(c\right)\right),\\
			\bm{f}_{t,k}\left(\bm{x}, \bm{\varphi}\right) &= \sqrt{\bar{\alpha}_{t}}E\left(P_{k}\left(\bm{x}, \bm{\varphi}\right)\right)+\sqrt{1-\bar{\alpha}_{t}}\bm{\epsilon},
		\end{aligned}
	\end{equation}
	where following~\citep{ho2020denoising,rombach2022high}, $\alpha_{t}=1-\beta_{t}$, $\bar{\alpha}_{t}=\prod_{i}^{t}\alpha_{t}$ is defined to construct the noised feature for timestep $t$. The closer the predicted noise to the actual noise, the higher the matching score. For consistency in representation, the output of CLIP can also be considered as the matching score. Then, the following estimated probability and prediction for a 3D point cloud sample under a single projection angle can be attained:
	\begin{equation}
		p_{\bm{\theta},K}(c\vert \bm{x},\bm{\varphi}) = \frac{{M\! S}_{\bm{\theta}, K}(\bm{x}, \bm{\varphi}, c)}{\sum_{j}{M\! S}_{\bm{\theta}, K}(\bm{x}, \bm{\varphi}, j)}\\
	\end{equation}
	\begin{equation}
		\hat{y}=\underset{c}{\arg\max}\ p_{\bm{\theta},K}(c\vert \bm{x},\bm{\varphi}).
	\end{equation}
	More details about the constructed semantic descriptions for different projection types can be found in Section~\ref{sec:Ex_diff}.

	\subsection{Angle Selection}
	\label{AS}
	To enhance the adaptability of the text-image matching framework to unaligned 3D point cloud data, the appropriate projection angles are indispensable. The choice of projection angles can be either predefined fixed angles, derived from the information of the 3D object, or refined according to preliminary matching scores.
	
	\subsubsection{Pre-defined Fixed Angles}
	
\begin{figure}[h]
		\centering
		\includegraphics[width=0.6\textwidth]{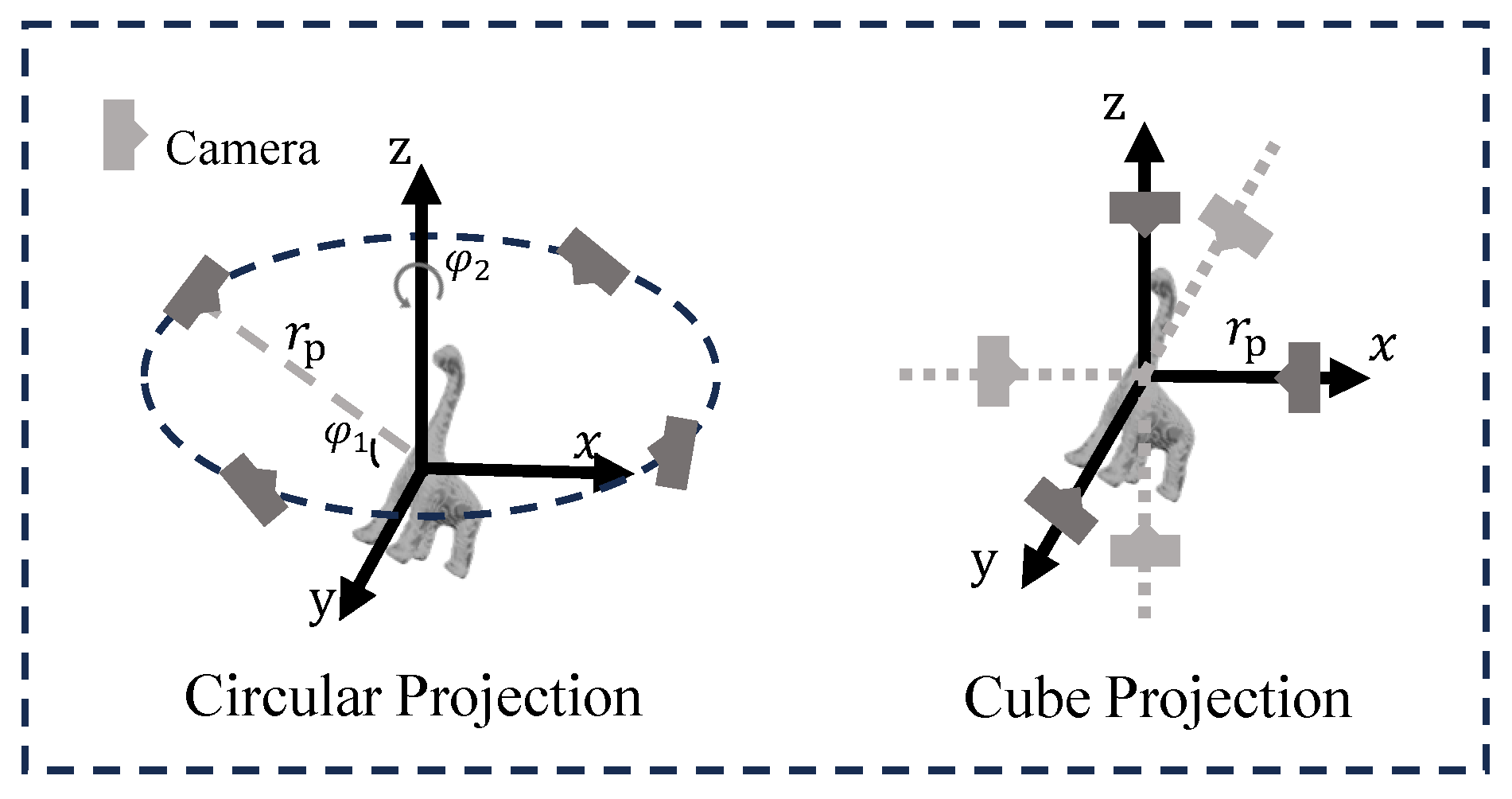}
		\caption{Basic Projection Methods}
		\label{fig:proj_sotas}
	\end{figure}

	Besides pre-selected single angles, existing pre-training-based 3D zero-shot classification methods only adopt circular and cube projections. Concerning circular projection, the camera is positioned obliquely above the object, which is adjusted by an angle $\varphi_{1}$ measuring the camera's inclination with respect to the x-y plane. Multi-view images are obtained by changing the azimuthal angle $\varphi_{2}$ of the camera's projection on the x-y plane relative to the x-axis. On the other hand, cube projection places the camera along the six directions of the three-dimensional coordinate axis. All cameras are oriented towards the center of the object and at a distance $r_p$ from the object. Both circular and cube projections keep the object pose unchanged, only adjusting the camera position, as shown in Fig.~\ref{fig:proj_sotas}. In this context, these two methods can achieve suitable views for open-pose 3D objects by fine-tuning the hyperparameters.
	
	\subsubsection{Iterative Angle Refinement Mechanism (IARM)}
	\label{sec:iarm}
	To enhance the identification of optimal projection angles, we propose an advanced method that integrates information from 3D objects and text-image matching. In instances where improper projection angles result in significantly low matching scores for specific classes, a high prediction probability does not guarantee a reasonable outcome. Thus, we introduce an iterative angle refinement module utilizing the projection angles $\bm{\varphi}_{c}=\left[\varphi_{c,1},\varphi_{c,2}\right]$, aimed at maximizing the text-image matching scores for each class $c$. This proposed module is expected to yield a more reasonable estimation of probability as depicted below:
	\begin{equation}
		p^{*}_{\bm{\theta},K}(c\vert \bm{x}) = \frac{{M\! S}_{\bm{\theta}, K}(\bm{x}, \bm{\varphi}_{c}, c)}{\sum_{j}{M\! S}_{\bm{\theta}, K}(\bm{x}, \bm{\varphi}_{j}, j)}\\
	\end{equation}
	
	\begin{equation}
		\bm{\varphi}_{c} = \underset{\bm{\varphi}}{\arg\max}\  {M\! S}_{\bm{\theta}, K}(\bm{x}, \bm{\varphi}, c).
	\end{equation}
	
	Assuming both the projection and text-image matching processes are differentiable, we can ascertain the projection angles yielding the highest matching score for each class through the gradient descent algorithm. Starting with an initial projection angle vector $\bm{\varphi}^{(0)}$, the entire iterative optimization process, incorporating a varying scaling factor $\left[\bm{\eta}_{r}\right]$, is outlined as follows:
	\begin{equation}
		\label{eq:update}
		\begin{aligned}
			\bm{\varphi}_{c}^{(0)} &= \bm{\varphi}^{(0)} \\
			\bm{\varphi}_{c}^{(r+1)} &= \bm{\varphi}_{c}^{(r)} + \bm{\eta}_{r}\cdot {sign}\left(\frac{\partial {M\! S}_{\bm{\theta}, K}(\bm{x}, \bm{\varphi}_{c}^{(r)}, c)}{\partial \bm{\varphi}_{c}^{(r)}}\right).
		\end{aligned}
	\end{equation}
	
	Upon completion of $R$ optimization steps, we utilize the estimated affine angle of each category $\widehat{\bm{\varphi}}_{c} = \bm{\varphi}_{c}^{(R)}$ to formulate the final classification prediction for the 3D point cloud samples, as outlined below:
	\begin{equation}
		\label{eq:pre}
		\widehat{y^{*}}=\underset{c}{\arg\max}\ \frac{{M\! S}_{\bm{\theta}, K}(\bm{x}, \widehat{\bm{\varphi}}_{c}, c)}{\sum_{j}{M\! S}_{\bm{\theta}, K}(\bm{x}, \widehat{\bm{\varphi}}_{j}, j)}.
	\end{equation}
	
	The prediction process aims to minimize the global risk associated with decision-making. Specifically, the confidence assigned to the angle most conducive to the correct classification of an individual object should surpass the confidences of other classes across all angles.
	
	In practice, the distribution of the matching score computed by the pre-training model often exhibits non-smoothness within the projection angle space. Additionally, certain semantic image matching models, such as the Stable Diffusion utilized in this study, entail a considerable number of parameters, rendering optimization impractical and computationally expensive in such scenarios.
	
	To streamline this optimization process, we propose a succinct strategy. Initially, we normalize and translate the sample coordinates to the coordinate origin to derive the covariance matrix $\bm{\Sigma}$:
	\begin{equation}
		\bm{\Sigma} = (\bm{x}^T \cdot \bm{x} ) / N, \quad \bm{\Sigma} \in \mathbb{R}^{3 \times 3},
		\label{eq:cov}
	\end{equation}
	where $\bm{x}$ represents the coordinates and $N$ denotes the number of vertices in a sample. By performing eigenvalue decomposition on this matrix $\bm{\Sigma}$, we can derive the three eigenvectors: $\bm{e_1}, \bm{e_2}, \bm{e_3} \in \mathbb{R}^{1\times3}$. Subsequently, we establish a new coordinate system for the sample, yielding the transformed coordinates as follows:
	\begin{equation}
		\bm{x'} = \bm{x}\cdot 
		\left[ \bm{e_1},\bm{e_2},\bm{e_3} \right].
		\label{eq:coor}
	\end{equation}

	\begin{algorithm}[hp]
		\caption{IARM on Diffusion Classifier}
		\KwData{\small{3D objects with open-pose: $\bm{x}$ \\
				\qquad \quad Semantic description for each class: $S(c)$ \\
				\qquad \quad Pre-training diffusion framework: $\bm{\epsilon _{\theta}}(), E()$ \\
				\qquad \quad Project function for each style: $P_{k}()$ \\
				\qquad \quad Number of iterations: $R$  \\  
				\qquad \quad Class number: $C$  \\  
				\qquad \quad Refine scales: $\left[\bm{\eta_{r}}\right]$ \\
				
				\qquad \quad Distance from the camera to the object: $r_p$ \\
		}}
		\begin{algorithmic}[1]
			\STATE Calculate the covariance matrix $\bm{\Sigma}$ through Eq.~\ref{eq:cov} and calculate its eigenvectors $\bm{e_1},\bm{e_2},\bm{e_3}$\\
			\STATE Establish a new coordinate system and obtain the coordinates $\bm{x'}$ through Eq.~\ref{eq:coor}\\
			\FOR{$c$ $\mathbf{in}$ range($C$)} 
			\STATE Initialize the projection angle $\bm{\varphi}_{c}^{(0)}$ = 0
			\FOR{ $r$ $\mathbf{in}$ range($R$)} 
			\STATE Sample random Gaussian noises $\bm{\epsilon}$ and calculate the matching score ${M\! S}_{\bm{\theta}, K}(\bm{x'}, \bm{\varphi}_{c}^{(r)}, c)$ through Eq.~\ref{eq:MD}
			\STATE Calculate the partial derivative of  matching score and update the $\bm{\varphi}_{c}^{(r)}$ through Eq.~\ref{eq:update}
			\ENDFOR
			\STATE Calculate  the matching score ${M\! S}_{\bm{\theta}, K}(\bm{x'}, \bm{\varphi}_{c}^{(R)}, c)$ through Eq.~\ref{eq:MD}
			\ENDFOR
			\STATE Obtain the final prediction $\widehat{y^{*}}$ through Eq.~\ref{eq:pre}
			\STATE \textbf{return} predicted label $\widehat{y^{*}}$
		\end{algorithmic}
		\label{alg}
	\end{algorithm}

	According to principal component analysis, fixing the camera on the z-axis in the new coordinates ensures the largest variance in each dimension of the projected image. This perspective may result in capturing more classification information. Furthermore, this dimensional reduction operation reduces the projection angles variable from two to one dimension (only the azimuthal angle $\varphi_{c,2}$). Instead of directly applying this rotation angle to the 3D data, it can be equivalently applied to the projected image. Following this initial adjustment of the projection angle, we can proceed to refine the angle through the previously introduced mechanism within a more simplified single-dimension case. The detailed steps for the proposed angle refinement strategy are outlined in Algorithm~\ref{alg}.

	\section{Experiment}
	\label{secE}
	\subsection{Experiment Setting}
	
	\noindent\textbf{Evaluation Metric.} Following the SOTA approaches~\citep{cheraghian2022zero,Hao2023ContrastiveGN,qi2023recon}, we utilize top-1 accuracy and employ mAcc to calculate the average accuracy across all categories, thereby providing a comprehensive reflection of the model's classification performance across different categories. Given that the pre-training-based methods discussed in this section do not necessitate training on native 3D datasets, there are no designated "seen" classes. To ensure a fair and uniform comparison, we exclusively evaluate the performance of conventional zero-shot classification among unseen classes.

	\noindent\textbf{Implementation Details.} We conduct our end-to-end inference process on a single RTX3090 card. The Stable Diffusion model~\citep{rombach2022high} is employed as the 2D pre-training model to predict the noise. Following Stable Diffusion, we set the seed to 42. For the hyperparameters of the diffusion classifier, we empirically tune the time steps and trials to 600 and 30, respectively. Additionally, we adjust the projected camera distance $r_p$ to 2.2 and the angle refinement parameters $R$ and $\left[\bm{\eta_{r}}\right]$ to 10 and $\left[20,18,16,...,2\right]$, respectively. Regarding projection, we utilize single-point light source projection from PyTorch3D~\citep{ravi2020pytorch3d} for mesh samples and voxel projection for point-cloud samples.
	
	\subsection{Analysis on Open-Pose 3D Zero-Shot Classification}
	\subsubsection{Comparison to SOTAs}
	\label{cts}
	
	We evaluate five recent SOTA methods: PointCLIP~\citep{ZhangGZLM0QG022}, Ulip~\citep{xue2023ulip}, ReconCLIP~\citep{qi2023recon}, CLIP2Point~\citep{huang2023clip2point}, and PointCLIPv2~\citep{Zhu2022PointCLIPV2}, on our open-pose benchmark McGill$^{\ddagger}$ and ModelNet10$^{\ddagger}$ for 3D zero-shot classification. All reproduction codes and pre-trained models are obtained from the official GitHub repository of the respective papers. The current state-of-the-art methods employ various pre-training models for zero-shot classification, such as GPT, CLIP, etc. Detailed information and results are provided in Table~\ref{tab:com_op}. ``Ours-CLIP" and ``Ours-Diffusion" indicate that we utilize CLIP and Diffusion as the pre-training text-image matching models in our pipeline (see Figure~\ref{fig:pipeline}), respectively.
	
	\begin{table}[h]
		\resizebox{0.99\textwidth}{!}{
			\begin{tabular}{cccccccccc}
				\bottomrule
				& Venue   & \multicolumn{4}{c}{Pre-training} & \multicolumn{2}{c}{McGill$^{\ddagger}$}                & \multicolumn{2}{c}{ModelNet10$^{\ddagger}$}            \\
				&         & GPT  & CLIP  & Diffusion & 3D-pm & Acc                 & mAcc                & ACC                 & mACC                \\ \hline
				PointCLIP      & CVPR'22 &      &  \checkmark     &           &       & 12.2                & 13.3                & 17.7                & 16.4                \\
				Ulip           & CVPR'23 &      &   \checkmark     &           &     \checkmark   & 14.8                & 16.1                & 14.4                & 13.8                \\
				ReconCLIP      & ICML'23 &      &    \checkmark    &           &    \checkmark    & 15.7                & 17.3                & 15.6                & 14.3                \\
				CLIP2Point     & ICCV'23 &      &    \checkmark    &           &      \checkmark  & 14.8                & 17.4                & 19.7                & 18.1                \\
				PointCLIPv2    & ICCV'23 &   \checkmark    &    \checkmark    &           &       & 27.8                & 28.9                & 19.9                & 18.2                \\ 
				\rowcolor{gray!20}  Ours-CLIP      & -       &      &    \checkmark    &           &       & \underline{ \textit{31.3}} & \underline{\textit{34.6}} & \textbf{26.3}       & \textbf{24.2}       \\
				\rowcolor{gray!20} Ours-Diffusion & -       &      &       &       \checkmark     &       & \textbf{39.1}       & \textbf{44.7}       & \underline{ \textit{22.6}} & \underline{\textit{21.7}} \\ \bottomrule
			\end{tabular}
		}
		
		\caption{Comparison to Current SOTAs on the Open-Pose 3D Zero-Shot Classification. 3D-pm stands for the \textbf{3D} \textbf{p}re-training \textbf{m}odel.}
		\label{tab:com_op}
	\end{table}
	
	\noindent\textbf{Results on the Open-Pose 3D Zero-Shot Benchmark McGill$^{\ddagger}$.} We present the results of our method on the McGill$^{\ddagger}$ benchmark in Table~\ref{tab:com_op}. Based on our pipeline, both Ours-CLIP and Ours-Diffusion exhibit compelling performance. Particularly, Ours-Diffusion demonstrates notable improvements of 11.3\% and 15.8\% on Accuracy (Acc) and mean Accuracy (mAcc), respectively. Furthermore, our approach relies solely on a single pre-training model (CLIP or Diffusion), rendering it a concise solution for initial open-pose 3D zero-shot classification.

	\noindent\textbf{Results on the Open-Pose 3D Zero-Shot Benchmark  ModelNet10$^{\ddagger}$.}
	As indicated in Table~\ref{tab:com_op}, our method also surpasses SOTAs with improvements of 6.4\% and 6.0\% on Accuracy (Acc) and mean Accuracy (mAcc), respectively. Unlike the results on McGill$^{\ddagger}$, the performance of Ours-Diffusion on ModelNet10$^{\ddagger}$ is lower than that of Ours-CLIP. We observe that CLIP exhibits particularly high accuracy in the 'toilet' and 'monitor' categories, whereas Diffusion demonstrates more evenly distributed performance across all categories. Due to the smaller number of classes, CLIP achieves better results on ModelNet10$^\ddagger$. Since CLIP utilizes feature similarity for measurement while Diffusion adopts noise MSE distance, it is challenging to directly compare or ensemble these two methods. In future work, we will explore methods to effectively combine CLIP and Diffusion to yield robust inference results. Notably, samples in these datasets tend to be boxy and symmetrical, with minimal variance between some classes, thus posing a challenging task in the open-pose scenario.
		
	\subsubsection{Analysis on Views Selection}
	
	\begin{table}[bp]
		\centering
		\resizebox{0.99\textwidth}{!}{
			\begin{tabular}{c|cccccccccccccc|cc}
				\bottomrule
				&Ant&Bird& Crab &Dino. &Dolp. &Fish&Hand &Octo.&Plier &Quad. &Snake &Spec. &Spider &Teddy & Acc & mAcc \\ \hline
				Top View& 0.0&28.6&0.0&0.0&0.0&0.0&0.0&0.0&42.9&0.0&0.0&\textbf{11.1}&9.1&0.0&6.1&6.6 \\
				Cube     &  0.0&57.1&0.0&0.0&\textbf{100.0}&0.0&14.3&25.0&85.7&0.0&66.7&0.0&18.2&0.0&21.7   &26.2\\
				Circular   & 0.0&57.1&0.0&0.0&75.0&\textbf{12.5}&57.1&\textbf{37.5}&\textbf{100.0}&9.1&55.6&0.0&9.1&0.0&25.2&29.5     \\
				\rowcolor{gray!20} Ours-IARM      & 0.0 &\textbf{71.4} &\textbf{10.0} &0.0 &\textbf{100.0} &\textbf{12.5} &\textbf{71.4} &\textbf{37.5} &71.4 &\textbf{27.3} &\textbf{77.8} &0.0 &\textbf{45.5} &\textbf{28.6} &\textbf{35.7} &\textbf{39.5}      \\ 
				\bottomrule
			\end{tabular}
		}
		\caption{Projection Angle Update on the McGill$^{\ddagger}$. In order to reduce the length of the table, we take abbreviations for some category names: Dino. vs Dinosaur, Dolp. vs Dolphin, Octo. vs Octopus, Quad. vs Quadruple, Spect. vs Spectacle.}
		\label{tab:angle}
	\end{table}
	
\begin{figure}[h]
		\centering
		\includegraphics[width=0.6\textwidth]{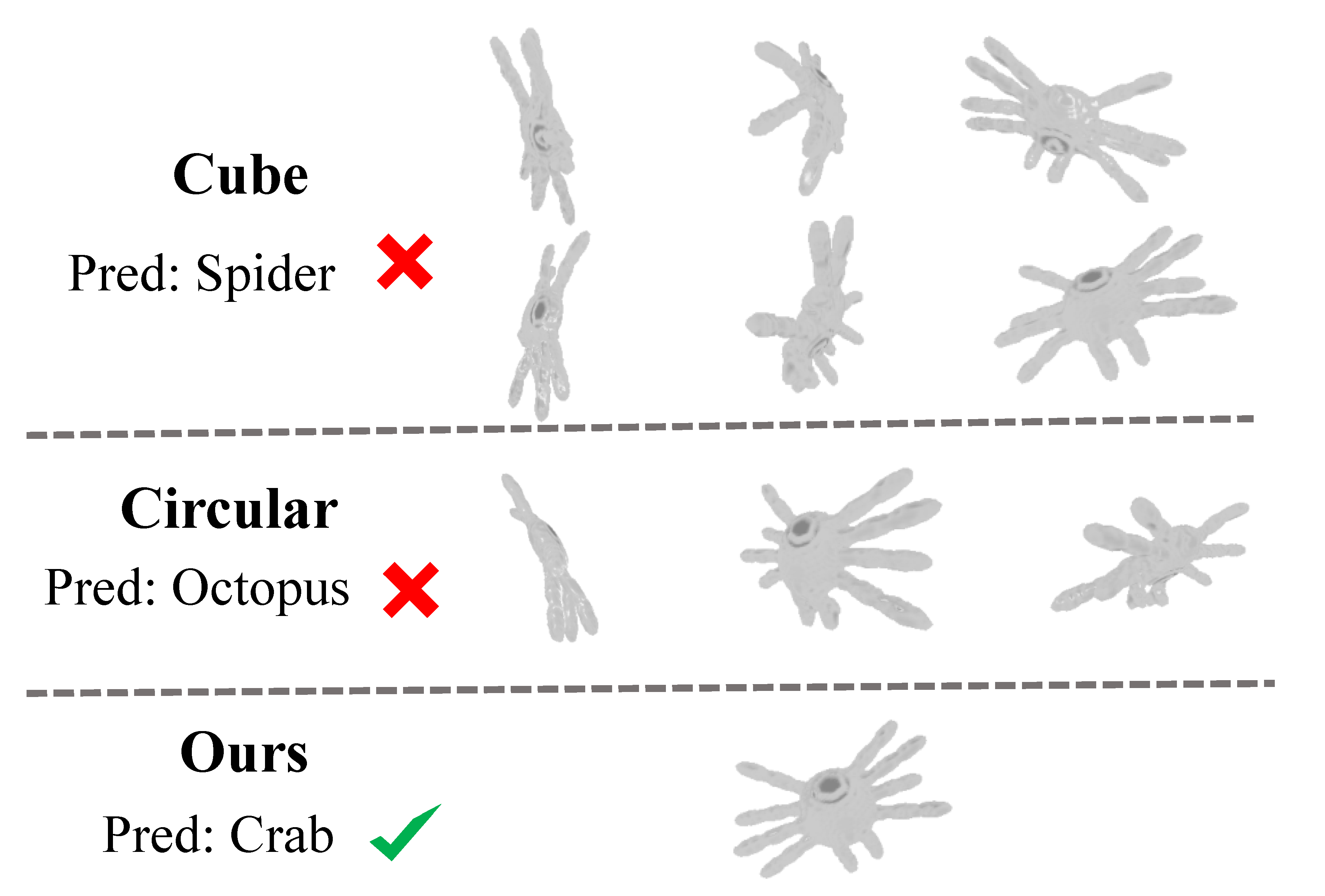}
		\caption{Final Views with the Corresponding Prediction.}
		\label{fig:view}
	\end{figure}

	We compare our Iterative Angle Refinement Mechanism (IARM) with commonly used projection methods, namely cube and circular views. The visualization results of these methods for a single sample, including angles, projections, and final predictions, are depicted in Figure~\ref{fig:view}. Additionally, we incorporate a fixed single-view approach utilizing only the top view for further context. The results, detailed in Table~\ref{tab:angle}, reveal significant disparities. In the open-pose scenario, the fixed single-view perspective yields notably poor results, with an accuracy of merely 6.6\%. While the cube and circular methods, as multi-view ensemble approaches, do show improvement over a single-view perspective, their performance is still hindered by the inherent randomness in open poses. In contrast, our angle refinement mechanism offers a more advantageous approach to selecting views conducive to classification. It leads to substantial gains, as evidenced by our method showing 10.5\% and 10.0\% improvements in Accuracy (Acc) and mean Accuracy (mAcc), respectively, on the open-pose McGill$^{\ddagger}$ benchmark.

	\subsection{Analysis on Bridging Pre-training Diffusion}
	\label{sec:Ex_diff}
	Unlike current SOTAs, we are the first to utilize 2D pre-training with diffusion as the 3D zero-shot classifier instead of CLIP. In the preceding section, we presented its final performance in the open-pose setting. Given the complexity inherent in the open-pose setting, we validate its performance on the generic aligned-pose ModelNet10 dataset to provide insights into its potential in detail.

	\subsubsection{Comparison on Prompts}
	First, we investigate the influence of semantic descriptions on the matching of various styled projections. We set the trials and camera distance $r_p$ to 30 and 2.2, respectively. We design multiple prompts for three style images: Render Image (Render I.), Depth Image (Depth I.), and Edge Image (Edge I.), and report the corresponding mAcc results on ModelNet10 in Table~\ref{tab:prompt}. Specifically, ``one line-drawn [$n_c$]" is the optimal prompt for depth images, ``one model of [$n_c$] in linear composition" is the best prompt for render images, while ``one edge map of one standalone [$n_c$]" is the most effective prompt for edge images. Moreover, the edge images exhibit the best results among the three types of images.
	
	\begin{table}[h]
		\centering
		\resizebox{0.8\textwidth}{!}{
			\begin{tabular}{cccc}
				\bottomrule
				Prompts                                                           & Render   I.& Depth I. &  Edge I.  \\ \hline
				\textit{one model of [$n_c$] }                                  &69.3&58.8&72.3 \\
				\textit{one line-drawn [$n_c$] }                                &69.9&\textbf{59.8}&72.0  \\
				\textit{one photo of one [$n_c$] }                              &69.6&53.2&73.7  \\
				\textit{one photo of one standalone [$n_c$] }                   &63.4&51.6&71.1  \\
				\textit{one depth map of one standalone [$n_c$] }               &67.2&48.5&77.4\\
				\textit{one edge map of one standalone [$n_c$] }               &66.1&51.9&\textbf{77.6}\\
				\textit{one render image of one standalone white  [$n_c$] }     &68.0&55.6&66.3\\
				\textit{one sketch photo of one standalone white [$n_c$] }      &65.8&53.0&74.7 \\
				\textit{one model of [$n_c$] in linear composition }            &\textbf{73.5}&59.5&73.5 \\
				\textit{one photo of one [$n_{c}$] in linear composition}         &73.1&59.7&73.7 \\
				\bottomrule
			\end{tabular}
		}
		\caption{Prompts for Multiple Style Images. \textit{[$n_{c}$]} stands for the name text of each category.}
		\label{tab:prompt}
	\end{table}
	
	\subsubsection{Comparison to the CLIP} 
	Subsequently, we compare the effectiveness of the diffusion classifier with that of the CLIP classifier on ModelNet10. To ensure a fair comparison, we only exchange the classifier while keeping all other conditions the same. The comparison results in terms of mAcc are reported in Table~\ref{tab:llm}. For CLIP, we utilize four common pre-training model structures: CLIP-VIT-B\textbackslash16~\citep{dosovitskiy2020image}, CLIP-VIT-B\textbackslash32~\citep{dosovitskiy2020image}, CLIP-ResNet50~\citep{he2016deep}, and CLIP-ResNet101~\citep{he2016deep}. Overall, CLIP-VIT-B\textbackslash16 yields the best performance among these CLIP models. However, the diffusion classifier demonstrates more powerful effectiveness, particularly when the trial is set to 30 (Tr.30).
	
	It is worth noting that the diffusion classifier is considerably slower than the CLIP classifier, primarily due to its more complex computational process. Theoretically, selecting a larger number of trials and time steps in diffusion could potentially lead to further enhancements in performance. Therefore, opting for the diffusion method entails a trade-off, wherein significant computational resources are required to achieve performance gains.
	
	\begin{table}[h]
		\centering
		\resizebox{0.7\textwidth}{!}{
			\begin{tabular}{ccccccc}
				\bottomrule
				& Render I.& Depth I. & Edge I. & Avg. & Times (s) \\ \hline
				CLIP-VIT-B\textbackslash16& 54.7 & 59.7 & 34.2 & 49.5 & \textbf{0.025}\\
				CLIP-VIT-B\textbackslash32& 49.6 & 52.7 & 38.0 & 46.8 & 0.028\\
				CLIP-ResNet50& 45.2 & 41.9 & 37.5 & 41.5 & 0.037\\
				CLIP-ResNet101& 48.4 &54.3 & 40.8 & 47.8 & 0.042\\ \hline
				Diffusion-Tr.10& 66.2 & 53.5 & 70.4 & 63.4 & 8.771\\
				Diffusion-Tr.20& 71.6 & 58.6 & 74.3 & 68.2 & 17.461\\
				Diffusion-Tr.30& \textbf{73.5} & \textbf{59.8} & \textbf{77.6} & \textbf{70.3} &26.078 \\ 
				\bottomrule
			\end{tabular}
		}
		\caption{Comparison to the CLIP. Times denotes the averaged inference time of a single projected image.}
		\label{tab:llm}
	\end{table}
	
	\subsubsection{Ablation on Projection Styles} 
	In the ablation studies of the three style images (Render, Depth, Edge) for both the CLIP and Diffusion models on the ModelNet10 dataset, the results are reported in Tables~\ref{tab:llm} and \ref{tab:img_style}. The CLIP model exhibits the best performance for depth images and the worst for edge images, whereas the diffusion model performs inversely. Additionally, we explore the ensemble of these styles to achieve better performance. Specifically, as introduced in Section~\ref{sec:DC}, the ensemble involves taking the average over projection styles when calculating the matching score. From Table~\ref{tab:img_style}, the CLIP model achieves better mAcc results by combining render and depth images. Conversely, the Diffusion model demonstrates the best performance by combining render and edge images.
	
	\begin{table}[h]
		\centering
		\resizebox{0.6\textwidth}{!}{
			\begin{tabular}{cccccc}
				\bottomrule
				& Render  & Depth  & Edge      &mACC  \\ \hline
				CLIP-VIT-B\textbackslash16& \checkmark & \checkmark&             & 61.2\\ 
				CLIP-VIT-B\textbackslash16&            & \checkmark& \checkmark  & 53.6\\ 
				CLIP-VIT-B\textbackslash16& \checkmark &           & \checkmark  & 50.2\\ 
				CLIP-VIT-B\textbackslash16& \checkmark & \checkmark& \checkmark  & 53.6\\ \hline
				Diffusion-Tr.30            & \checkmark & \checkmark&             &73.2 \\
				Diffusion-Tr.30            &            & \checkmark& \checkmark  &79.4\\
				Diffusion-Tr.30            & \checkmark &           & \checkmark  &\textbf{81.7} \\ 
				Diffusion-Tr.30            & \checkmark & \checkmark& \checkmark  &79.4 \\ 
				\bottomrule
			\end{tabular}
		}
		\caption{Ablation on Projection Styles}
		\label{tab:img_style}
	\end{table}
	
	\section{Challenges}
	The open-pose setting poses greater difficulty for 3D zero-shot classification and holds more practical significance. In this context, we make the first attempt to introduce the 3D open-pose zero-shot classification task and provide one effective method as the baseline for subsequent studies. Clearly, this task presents numerous challenges that warrant further research and investigation. In this section, we focus on exploring several key challenges and potential solutions.

	\textbf{Insufficient samples for 3D seen classes.} The scarcity of samples for 3D seen classes presents a significant obstacle to training accurate classification models. Unlike 2D data, the scarcity of 3D data arises from higher acquisition costs, increased processing and storage requirements, limited access channels, and difficulties in annotation. To overcome this challenge, researchers may explore data augmentation techniques such as scaling and translation to generate synthetic samples and enrich the training dataset. Additionally, leveraging text-to-3D generation models~\citep{lin2023magic3d, chen2023fantasia3d} or image-to-3D generation models~\citep{liu2023zero} to expand 3D data holds promising prospects.
	
	\textbf{Viewpoint similarity among the classes.}  The presence of viewpoint similarity among different classes complicates the 3D zero-shot classification process, particularly for projection-based methods~\citep{huang2023clip2point, Zhu2022PointCLIPV2, xue2023ulip, wang2023transferring}. Instances like tables and beds can appear remarkably similar from certain viewpoints, a common occurrence in the open-pose setting. One possible solution is to incorporate viewpoint augmentation during training, exposing the model to diverse viewpoints of the same object class to improve its robustness against viewpoint variations. For methods that do not require additional training, iterating multiple times may help determine the optimal viewpoint.

	\textbf{Attribute relationship between seen and unseen classes.} Establishing an effective relationship between seen and unseen classes is crucial for generalizing the classification model to unseen classes~\citep{zhou2023attribute,wang2023learning}. Transfer learning techniques, such as feature alignment and domain adaptation, can be explored to leverage knowledge from seen classes and transfer it to unseen classes. Furthermore, fundamental knowledge from disciplines such as physics and biology could be utilized to generalize the attributes of seen and unseen classes.
	
	\textbf{Bias in the distribution of pose data.} In training data for large models, there is typically a lower proportion of open-pose data compared to aligned-pose data~\citep{radford2021learning, rombach2022high, sohl2015deep}. The imbalance in the distribution of open-pose data in training datasets hinders the model's ability to learn representative features for open-pose classification. Addressing this challenge may involve collecting additional open-pose data or exploring techniques to balance the distribution of open-pose data in the training process.

	Our open-pose setting introduces additional complexities, such as variations in object orientations and viewpoints, which are not adequately addressed by existing 3D zero-shot classification models. By incorporating these considerations into the design and training of 3D large-scale models, there will be an enhancement in their adaptability and robustness for handling diverse 3D data in reality. Our work serves as an important clue, highlighting the critical need for open-pose scenarios to promote the development of 3D zero-shot learning.

	\section{Conclusion}
	\label{secC}
	This paper provides an overview of the current progress in 3D zero-shot classification and proposes a more challenging benchmark for 3D zero-shot classification, aiming to recognize unseen 3D objects with open poses. Correspondingly, we validate the effectiveness of different strategies and design a concise pipeline with a concise angle refinement mechanism to present the preliminary solution.  However, due to the significantly higher difficulty, our approach, being the first exploration in open-pose situations, does not achieve as remarkable results as in the aligned-pose case. Furthermore, we pioneer the exploration of knowledge transfer using pre-training Diffusion, broadening the scope of validation beyond existing CLIP-based methods. Finally, we also set out challenges and potential exploration strategies for 3D zero-shot classification in the open-pose setting.

	\bibliographystyle{elsarticle-harv} 
	\bibliography{ref}
	
\end{document}